\documentclass{article}

\usepackage{arxiv}
\usepackage[utf8]{inputenc} 
\usepackage[T1]{fontenc}    
\usepackage{hyperref}       
\usepackage{url}            
\usepackage{booktabs}       
\usepackage{amsfonts}       
\usepackage{nicefrac}       
\usepackage{microtype}      
\usepackage{lipsum}
\usepackage{graphicx}
\graphicspath{ {./images/} }

\usepackage{stmaryrd} 
\usepackage{amsmath,amssymb}

\usepackage{adjustbox}
\usepackage{caption}
\usepackage{subcaption}
\usepackage{multirow}
\usepackage{setspace}
\usepackage{graphicx}
\usepackage{amssymb}
\usepackage{amsmath,amssymb,amsfonts}
\usepackage{adjustbox}
\usepackage{algorithmic}
\usepackage{caption}
\usepackage{textcomp}
\usepackage{xcolor}
\usepackage{booktabs}
\usepackage[normalem]{ulem}
\usepackage{threeparttable}
\usepackage{hyperref}
\usepackage{xcolor}
\usepackage{enumitem}

\useunder{\uline}{\ul}{}

\def\BibTeX{{\rm B\kern-.05em{\sc i\kern-.025em b}\kern-.08em
		T\kern-.1667em\lower.7ex\hbox{E}\kern-.125emX}}

\usepackage{lineno}

\title{Autoencoders for Strategic Decision Support}

\author{
Sam~Verboven \\
Data Analytics Laboratory \\
Solvay Business School\\
Vrije Universiteit Brussel\\
Brussels, Belgium\\
\texttt{sam.verboven@vub.be}\\
\And
Jeroen-Berrevoets\\
Data Analytics Laboratory \\
Solvay Business School\\
Vrije Universiteit Brussel\\
Brussels, Belgium\\
\texttt{jeroen.berrevoets@vub.be}
\And
Chris-Wuytens\\
Antwerp Management School \\
Universiteit Antwerpen \\
Antwerp, Belgium\\
\texttt{chris.wuytens@ams.ac.be}
\And
Bart-Baesens\\
Faculty of Business and Economics \\
Katholieke Universiteit Leuven\\
Leuven, Belgium\\
\texttt{bart.baesens@kuleuven.be}
\And
Wouter~Verbeke \\
Data Analytics Laboratory \\
Solvay Business School\\
Vrije Universiteit Brussel\\
Brussels, Belgium\\
\texttt{wouter.verbeke@vub.be}
}

\begin{document}
\maketitle
\begin{abstract}
In the majority of executive domains, a notion of normality is involved in most strategic decisions. However, few data-driven tools that support strategic decision-making are available. We introduce and extend the use of autoencoders to provide strategically relevant granular feedback.
A first experiment indicates that experts are inconsistent in their decision making, highlighting the need for strategic decision support.
Furthermore, using two large industry-provided human resources datasets, the proposed solution is evaluated in terms of ranking accuracy, synergy with human experts, and dimension-level feedback. This three-point scheme is validated using (a) synthetic data, (b) the perspective of data quality, (c) blind expert validation, and (d) transparent expert evaluation. 
Our study confirms several principal weaknesses of human decision-making and stresses the importance of synergy between a model and humans.
Moreover, unsupervised learning and in particular the autoencoder are shown to be valuable tools for strategic decision-making.
\end{abstract}

\section{Introduction}
\subsection{Problem Description}

Data-driven approaches, such as machine learning and artificial intelligence methods, perform best at well-defined tasks that are repetitive and have tractable short-term effects. 
As such, these methods are being adopted across industries to support, optimize and automate operational decision-making \cite{bertsimas2019predictive}.
Examples include the adoption of machine learning in credit scoring to optimize decisions to extend credit \cite{baesens2003using}, and in customer churn prediction to optimize customer relationship management \cite{verbeke2012new}.
Regardless of these successes in operational decision support, strategic decision-making is still rarely supported by learning-based systems \cite{lee2018understanding}.
A possible reason is that the synergy with supervised algorithmic methods is much less evident, as strategic decision-making often involves one-off decisions that are not well-defined (there are no labels) and have long-term effects. 

So in spite of technological progress, which has made operational task support such as churn prediction accessible to the average company, strategic decisions are still predominantly made without any learning-based grounding.
As such, the most important long-term decision-making in an organization is arguably the least supported by learning systems.
Hence, strategic decision-making has to date been guided by expert knowledge even though humans are known to be prone to various biases \cite{kahneman2013choices}, and managers are known to have preconceptions that lack objective grounding \cite{doukas2007acquisitions}. A lack of data-driven strategic decision support thus represents a large-impact problem across industries. 

\subsection{Solution Requirements} \label{subsec:requirements}

\begin{figure}[t] 
	\begin{center}
		\includegraphics[scale=0.75]{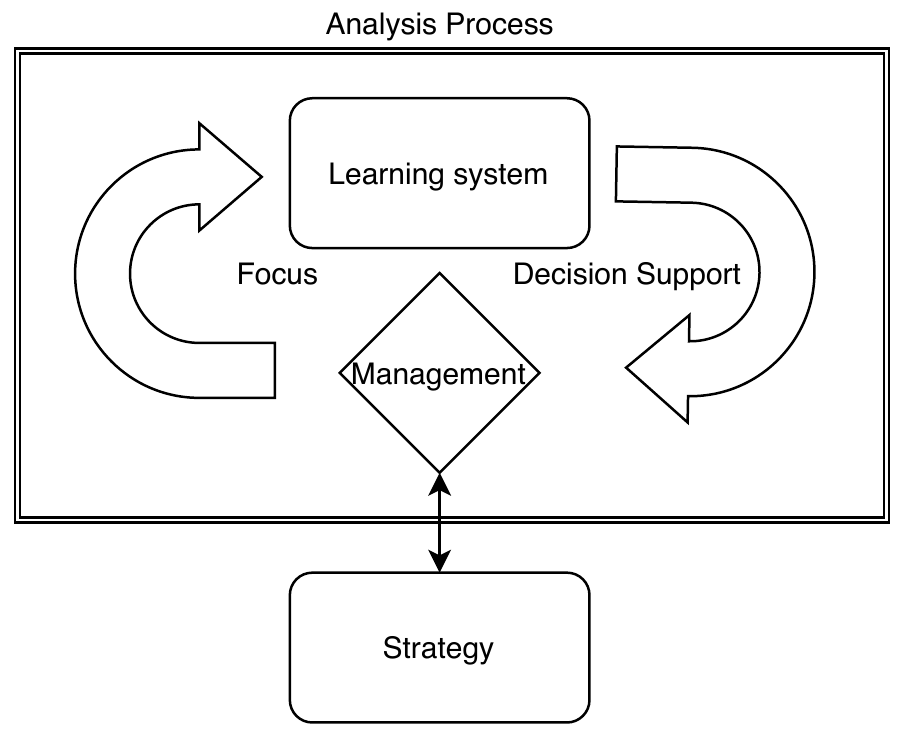}
		\caption{Conceptual diagram of the role of a learning system in decision support}
		\label{figdecisionsupport}
	\end{center}
	\vspace*{-6mm}
\end{figure}

A data-driven solution of this problem would entail a system capable of learning from data that provides management with detailed and actionable information, as conceptually displayed in Figure~\ref{figdecisionsupport}.
As labeled examples are generally unavailable in strategic decision-making, this information must be learned using unsupervised learning.

Currently, lacking a learning-based grounding, a manager matches an internalized representation of what is expected with reality, i.e., the definition of a correct remuneration of an employee is dependent on the market, and a revenue increase of one percent appears less positive if all competitors grow by ten percent. 
As such, a notion of normality is necessary for management to function and intelligently outline strategy \cite{liebl2010normality}. 

The field concerned with this notion of normality is called outlier or anomaly detection, the identification of unexpected or abnormal behavior \cite{aggarwal2001outlier}.
Nonetheless, outlier detection does not suffice to enable provision of detailed and actionable information.
Whereas in outlier detection the general objective is to select the subset of entities that differ the most from the other observations, here we aim to (1) learn for each entity in the dataset (2) whether, (3) how, and (4) to what extent an organization is different from relevant peers. In other words, to provide actionable strategic information, we require \textbf{granular feedback} on the outlyingness of each entity in the population. The objective in strategic decision-making, as stipulated by the above four technical requirements, significantly extends the much simpler objective in traditional outlier detection applications.

However, one cannot hope to solve this problem by only looking at the technical side of algorithms; the output of the solution is required to be actionable, interpretable, justifiable \cite{martens2011}, and ultimately accepted by the decision-makers.
We define these qualitative prerequisites as a requirement of (5) \textbf{synergy}.
This fifth requirement is often ignored but is key to ensuring the eventual adoption of the proposed decision support system. Past research has shown strong distrust and even dislike towards algorithmic decision support in the managerial domain \cite{lee2018understanding} and, ultimately, the management is responsible for the executive decisions.

\subsection{Solution} \label{subsec:introsolution}

\begin{figure}[t] 
	\begin{center}
		\includegraphics[scale=0.55]{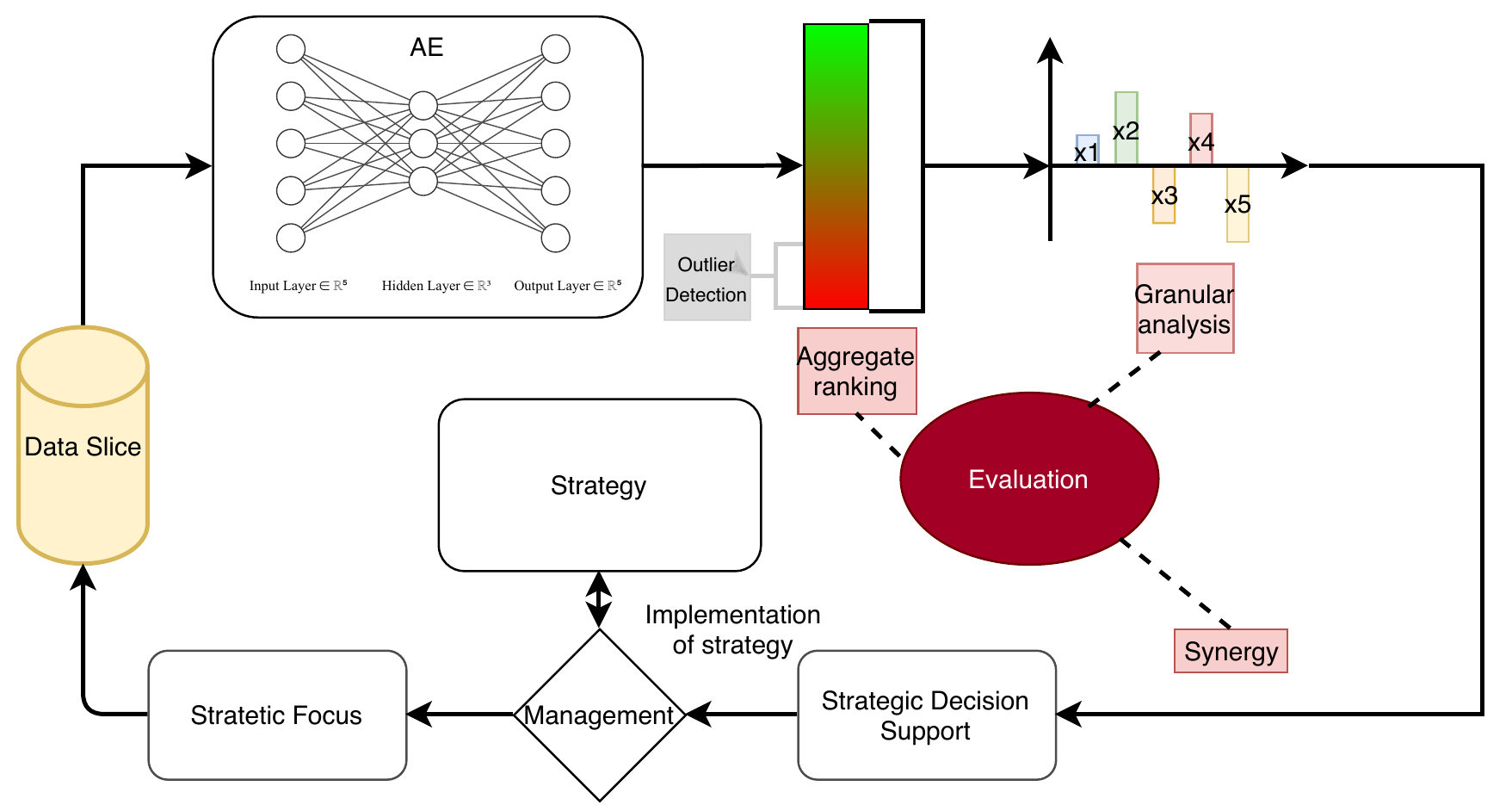}
		\caption{Comprehensive Diagram of the Extension of Data-driven Decision Support}
		\label{figdecisionsupport2}
	\end{center}
	\vspace*{-6mm}
\end{figure}

In this study, we introduce a framework for providing strategic data-driven decision support by utilizing an autoencoder (AE) neural network to provide a population-wide detailed analysis of the situation as-is \cite{smith1992}. 
Consequently, this framework facilitates provision of granular feedback based on data by explicitly scoring in terms of how, to what extent and in what sense an observation deviates from a learned normal state, i.e., from what is expected, given the particular context when comparing to a set of relevant peers. Such a diagnosis supports decision-making with the aim of arriving at an improved situation to-be by highlighting comparative advantages and disadvantages and by providing a sense of the size and importance of competitive differences. This allows determination of priorities and courses of action. Simply put, by means of comparison with a relevant benchmark, one may learn how to change for the better. 
The structure of the solution is visualized in Figure~\ref{figdecisionsupport2}. 

We introduce an extensive evaluation approach specifically designed to gauge the capacity to fulfill every single requirement for effective strategic decision support defined in Section~\ref{subsec:requirements}.
As 
Figure \ref{figdecisionsupport2} shows, this still involves an evaluation of not only the aggregate ranking but also elements of the granular analysis as well as synergy with management. The evaluation framework is discussed in full detail in Section~\ref{sec:experiments}.

For the purpose of the presented study, two proprietary datasets were obtained from a European HR services provider; these are sets of observations representing employees (D1) and employers (D2), including a selection of five and eleven dimensions of employees and employers, respectively. These datasets allow us to evaluate the use of the proposed approach to leverage unlabeled datasets for providing relevant input to the strategic decision-making process.

\subsection{Contributions}
In this article, we introduce and extend the use of autoencoders to assess the level of normality across a full set of observations for providing strategic decision support. We introduce and apply an assessment procedure to validate the proposed methodology using two HR datasets. We leverage data quality issues, expert opinion, expert validation and synthetic observations to demonstrate that the AE-based method does the following: 
\begin{itemize}
	\item Outperforms humans and other benchmark models;
	\item Offers granular dimension-level feedback, yielding extensive insights beyond the aggregate outlier scores; and
	\item Outputs information considered relevant and interpretable and is thus highly synergetic with human experts.
\end{itemize}

We present experimental results that validate (a) the business need for a data-driven diagnosis and (b) the adequacy of the proposed methodology in providing such decision support. The presented application of the proposed methodology is in the field of human resources management. However, the methodology is versatile and can be applied across management domains and at various levels of an organization by selecting an appropriate dataset relevant for the envisioned analysis.

The remainder of this paper is structured as follows. In the next section, the related literature is reviewed. Subsequently, in Section~\ref{sec:methodology} the proposed methodology is discussed. In Section~\ref{sec:problemvalidation}, the need for data-driven strategic decision support is experimentally demonstrated. Next, Section~\ref{sec:experiments} describes a series of experiments evaluating the effectiveness of the autoencoder as a solution. The implications of these experiments are reported and discussed in Section~\ref{sec:discussion}. Finally, conclusions and future research opportunities are presented in Section~\ref{sec:conclusion}.

\section{Related Work} \label{sec:related_work}
In this section, we first review strategies for handling label imperfections as labeled observations are not naturally available for decision support. Nevertheless, strategies based on human annotations could be helpful, but ultimately depend on the reliability of human decision-making.
Next, to characterize human decision making, relevant studies from cognitive psychology are highlighted.
Furthermore, the outlier detection literature is reviewed, as it relates closely to strategic decision support.
Finally, unsupervised learning for outlier detection is specifically examined to accommodate the absence of labels in the setting of this paper.

\subsection{Strategies for Label Imperfections}
\begin{figure}[t] 
	\begin{center}
		\includegraphics[scale=0.65]{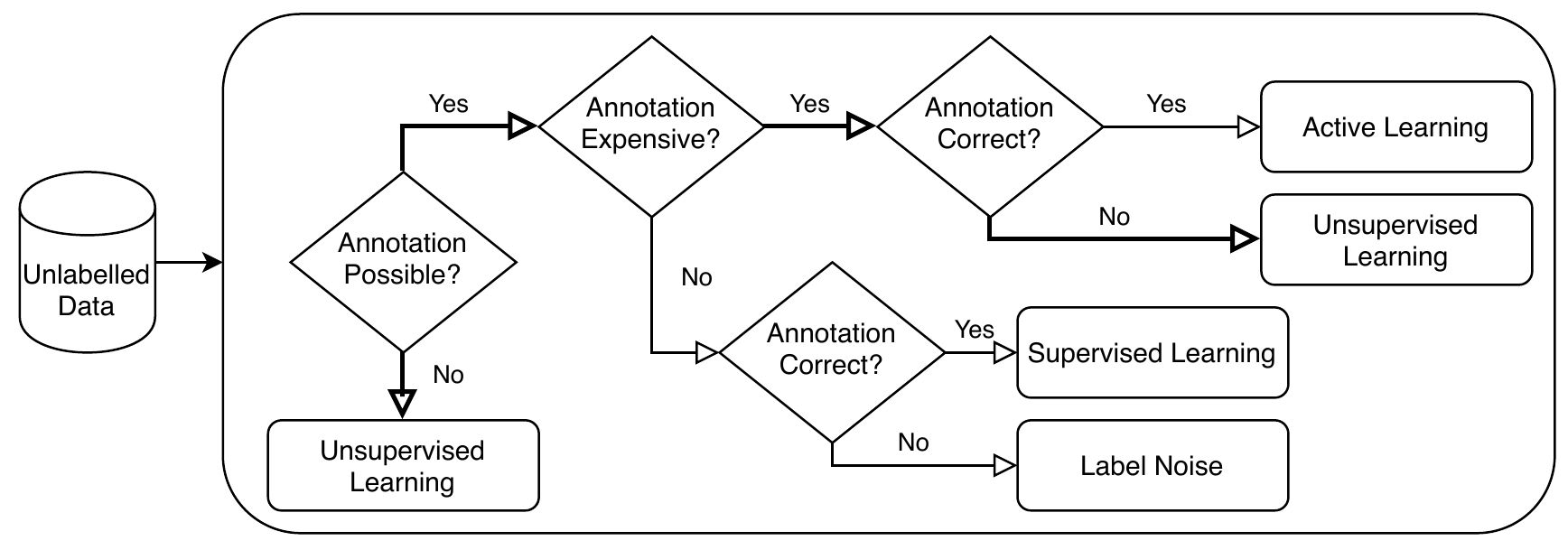}
		\caption{Label Imperfection}
		\label{tree}
	\end{center}
	\vspace*{-6mm}
\end{figure}
In the context of the research goal in this paper, no labeled data is naturally available.
Whether annotation-based strategies are possible or unsupervised learning is necessitated also hinges on human expertise.
A visual summary of this section is shown in Figure \ref{tree}.
Supervised learning methods require labeled observations, but labels obtained through human expert annotation are often expensive to obtain or incorrect. 

If annotation is inexpensive but incorrect, `label noise'-based learning techniques are an attractive approach \cite{natarajan2013learning}. Deep learning, specifically, has been shown to be able to deal with severe label noise \cite{rolnick2017deep}.

If annotation is expensive but of high quality, semi-supervised and active learning methods can be adopted.
In semi-supervised settings, a limited number of labels are used to aid training \cite{zhu2005semi}.
A limited set of human-assigned labels is leveraged in active learning \cite{abe2006outlier}. 
Here, the model optimizes queries to a human oracle to label a limited number of unlabeled instances. The use of human feedback allows aligning the inferred labels with the semantic understanding of the experts \cite{das2016incorporating}.

If annotation is unavailable, or expensive and incorrect, unsupervised learning can offer solace, as it relies solely on the internal structure of the data. For decision support, labels are always expensive, and almost always unavailable. 

\subsection{Elements of Cognitive Psychology}
Whether labels obtained through annotation are reliable thus relies on human expertise. The field of cognitive psychology helps us understand the limits of human feedback \cite{sterman1989modeling}.
Logical consistency across decisions, regardless of presentation, is a cornerstone of rational decision-making \cite{de2006frames, kahneman2013choices}. Over the years, an overwhelming amount of data has refuted this `description-invariance' hypothesis \cite{kahneman2013choices, de2006frames}. Moreover, humans' initial emotional or intuitive responses affect judgment as well \cite{de2006frames}. Scalability is affected by fatigue and boredom \cite{evans1987attentional}, as well as limited short-term memory \cite{kahneman1973attention}. Finally, when faced with complex information, humans fall back to simple heuristics \cite{kahneman2002representativeness, gigerenzer2011heuristic, van1998improving}.

Moreover, humans are not always aware of their performance. This notion broadly refers to term `meta-cognition', or knowing about knowing. Kruger and Dunning \cite{kruger1999unskilled} argue that the extent of humans' relative ignorance is often unknown to humans themselves. Furthermore, they have collected ample evidence of a particular embodiment of this issue, called the Dunning-Kruger effect \cite{kruger1999unskilled}. The latter \cite{kruger1999unskilled} is a cognitive bias observed in many social and intellectual domains, in which individuals who are incompetent at a particular task tend to overestimate their competence. It has been observed that the more incompetent individuals are, the larger the overestimation of their relative performance \cite{kruger1999unskilled}.

In summary, the impact of these limitations on decision support systems is threefold:
\begin{enumerate}[label=(\alph*)]
	\item Human cognition suffers from several flaws and biases and needs objective grounding.
	\item Human judgment of own expertise is imperfect, making humans inadequate judges of the need for algorithm-based decision support.
	\item If human judgment is too inconsistent or cannot cope with high dimensions, an unsupervised learning paradigm must be adopted
\end{enumerate}  
To assess the impact of these limitations on learning and developing a data-driven strategic decision support system, it is paramount to study human expertise.

\subsection{Outlier Detection}
Our study differs from traditional outlier detection applications, as we are interested in the distribution and behavior of the entire population, rather than in identifying a small subset of outlying observations. Outlier detection has been successfully applied in a plethora of fields, including fraud detection \cite{baesens2015fraud}, computer vision \cite{mahadevan2010anomaly}, network intrusion detection \cite{bhuyan2013network}, and medicine \cite{vanleemput2001}. The interest in outlier detection stems from the assumption that identification of outliers and their characteristics translates into actionable information \cite{chandola2009anomaly}. We extend this assumption and argue that common unsupervised methods may uncover information relevant to support decisions across the entire population. Note that in the absence of labeled observations, unsupervised methods allow the ranking of observations based on the level of outlyingness indicated by outlier scores.

\subsection{Unsupervised Outlier Detection}
Approaches to unsupervised outlier detection are mostly based on statistical reasoning, distances, or densities \cite{schubert2012evaluation}. The capacity of methods to accurately identify outliers varies across applications and depends on the dimensionality of the dataset, although some methods appear to be robust and generalize better than others \cite{campos16}.

Typically, a score is produced that can subsequently be used to rank entities based on how outlying they are.
The nature of such rankings produced by unsupervised outlier detection techniques is not yet well understood \cite{campos16, zimek2014}. This implies that every method inherently adopts its own implicit definition of what constitutes normality. The optimal definition varies across application domains \cite{campos16}. The autoencoder \cite{baldi2012autoencoders} is a method that combines strong performance with a possibility of granular feedback. Deep autoencoding architectures have achieved outstanding results in traditional outlier detection \cite{lu2017unsupervised, zhou2017anomaly, fan2018analytical}.

To evaluate unsupervised models, expert input, e.g., a set of observations labeled by an expert, can be used; alternatively, if labels are available (though unused by the unsupervised learning method), then a holdout test set can be used as in the evaluation of supervised models \cite{schubert2012evaluation}.
An evaluation based on expert input hinges on two critical assumptions:
\begin{enumerate}[label=(\roman*)]
	\item the expert's labeling is correct, and
	\item the expert's semantic understanding is relevant or desirable.
\end{enumerate}

Note here that labeling observations becomes exceedingly difficult if the dimensionality of the observations, i.e., the number of available dimensions, increases \cite{sakurada2014anomaly}.
This implies that the relevance of expert input is limited.

\section{Methodology} \label{sec:methodology}
In this section, we will discuss the autoencoder as well as two other state-of-the-art outlier detection methods, namely, the local outlier factor (LOF)\cite{breunig2000lof} and isolation forest (Iforest)\cite{liu2008isolation}. 

\subsection{Autoencoders for decision support}
The autoencoder facilitates an extension to decision support by offering granular and actionable information in addition to an overall outlier score and ranking. LOF and Iforest do not offer such granular feedback but will be used in the experiments to benchmark the outlier ranking obtained from the autoencoder.

\begin{figure}[h] 
	\begin{center}
		\includegraphics[scale=0.70]{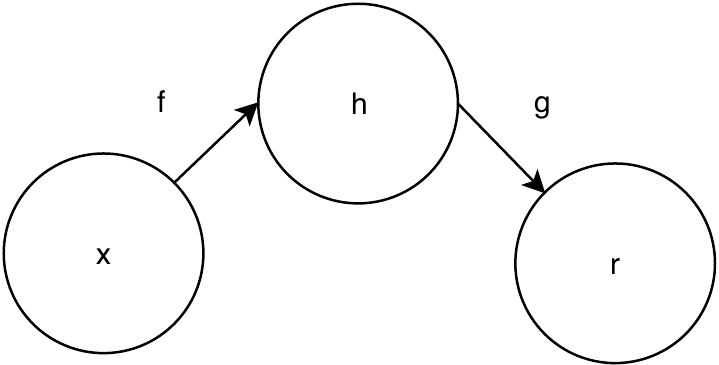}
		\caption{Autoencoder (adapted from \cite{goodfellow2016deep})}
		\label{fig1}
	\end{center}
	\vspace*{-6mm}
\end{figure}

Autoencoders are symmetric artificial neural networks trained with the objective of reconstructing their inputs, i.e., observations. A basic autoencoder (Figure~\ref{fig1}) maps an input vector $ \mathbf{x} \in \mathbb{R}^n $, where $n \in \mathbb{N}^+$ is the dimension of $\mathbf{x}$, to an output vector of an equal dimension, i.e., the reconstructed observation $ \mathbf{r} \in \mathbb{R}^n $.
An autoencoder essentially consists of two main components: (1) an encoder $f$ that maps $ \mathbf{x} $ to an internal representation $ \mathbf{h} \in \mathbb{R}^m $, where $m \in \mathbb{N}^+$, and (2) a decoder $g$ that maps $ \mathbf{h} $ to $ \mathbf{r} $.

Most applications of autoencoders aim to extract useful properties of the dataset through the internal representation $ \mathbf{h} $. Such applications include pre-training \cite{vincent2010stacked}, dimensionality reduction \cite{hinton2006reducing}, and vectorizing word representations \cite{ap2014autoencoder}. However, in outlier detection and by extension in decision support, we are primarily interested in the output $ \mathbf{r} $. More specifically, here we are interested in the similarity between $ \mathbf{r} $ and $ \mathbf{x} $ expressed by a loss function 
$\mathcal{L}(\mathbf{x}, g(f (\mathbf{x})))$. Generally, a loss function that penalizes the distance from $ \mathbf{r} $ to $ \mathbf{x} $ is selected, thereby defining the reconstruction error. By restricting the capacity of $ \mathbf{h} $, useful properties of the data may be learned \cite{goodfellow2016deep}. 
In an undercomplete autoencoder, the internal representation acts as a bottleneck since $ \mathbf{h} $ is of a lower dimension than $ \mathbf{x} $, i.e., $n>m$. Through this bottleneck, an incomplete reconstruction is forced since model capacity no longer suffices for an exact reconstruction. 
When training the autoencoder with the objective of minimizing the reconstruction loss, we implicitly favor the reconstruction of inputs that \emph{are closest to the data}. Hence, inputs that are the farthest from the learned reconstruction exhibit the largest errors. If more hidden layers are used in the autoencoder architecture, the capacity of the network increases, enabling it to construct a more complex hidden encoding of the data.

An undercomplete autoencoder combines multiple characteristics of an attractive solution to our problem:
\begin{itemize}
	\item The reconstruction errors can be interpreted as deviations for each individual dimension from the \emph{normal} or \emph{expected} state, and 
	\item The errors offer information about both the size and the direction of the deviation.
\end{itemize}

\subsection{Outlier ranking methods}\label{subsec:outlierrankmethods}
\subsubsection{Local Outlier Factor.}
The local outlier factor method (LOF) \cite{breunig2000lof} is a state-of-the-art unsupervised outlier detection algorithm \cite{goldstein2016comparative}. LOF is a density-based scheme in which an outlier score $LOF_k(p)$ is computed for each observation. 

The $k$ nearest neighbors $N_k(p)$ are determined for each observation $p$, where $k \in \mathbb{N}^+$. Afterwards, the local reachability density $lrd_k(p)$ for one observation $p$ is computed:
\begin{equation}\label{eq:lrd}
lrd_k(p) = \left (\frac{\sum\limits_{o\in N_k(p)}d_k(p,o)}{\left | N_k (p) \right |}  \right )^{-1},
\end{equation}
where $d_k$ is the reachability distance. In (\ref{eq:lrd}), the local reachability density is thus inversely proportional to the average reachability distance from $p$ to its $k$ neighbors. The reachability distance is almost always computed as the Euclidean distance \cite{goldstein2016comparative}. Intuitively, a larger distance between observations implies a lower density.

Given $lrd_k(p)$, $LOF_k(p)$ can be computed:
\begin{equation}
LOF_k(p) = \frac{\sum\limits_{o\in N_k(p)}\frac{lrd_k(o)}{lrd_k(p)}}{| N_k (p) |}.
\end{equation}
$LOF_k(p)$ is the average ratio of the lrd of $p$ to the lrds of its k neighbors.

The number of nearest neighbors being considered ($k$), and the distance measure for the reachability distance, i.e., Euclidean, are hyperparameters of the model (cf. Table~\ref{tab:hyperparameters} in the Appendix). Observations with a density that is substantially lower than those of their neighbors are considered outliers or anomalies.

As the average ratio between the densities of the observation and the neighborhood increases, so does $LOF_k(p)$. Hence, $LOF_k(p)$ being equal to one implies that $lrd_k$ of observation $p$ is on average equal to $lrd_k$ of its neighbors. A higher $LOF_k(p)$ indicates that $p$ lies, on average, in a lower-density area than those of its neighbors and can thus be considered to be more outlying. LOF outputs a score that can subsequently be used to rank observations from high to low level of outlyingness.

\subsubsection{Isolation Forest (Iforest).}
Isolation forest (Iforest) \cite{liu2008isolation} is a powerful outlier detection algorithm that extends decision tree and ensemble methods, such as random forests. Isolation implies ``\textit{the separating of an instance from the rest of the instances}'' \cite{liu2008isolation}. The key assumption behind Iforest is that anomalies are fewer and different and are thus more susceptible to \emph{isolation} when the input space is randomly segmented.

Compared to inlying observations, an outlying observation will on average require fewer splits of a decision tree that randomly partitions the input space, for the observation to be isolated from other observations. If a forest of such random trees collectively produces shorter \emph{path lengths} for some observations to be isolated, the latter are likely outliers.

The number of edges an observation $x$ traverses in an isolation tree from the root node to termination at an external node is denoted by $h(x)$. Moreover, a normalization factor $c(n)$ enables comparisons across different subsampling sizes. The Iforest method then calculates a score $s(x,n)$,
\begin{equation}
s(x,n) = 2^{-\frac{\mathbb{E}[h(x)]}{c(n)}},
\end{equation}
where $\mathbb{E}[h(x)]$ is the expectation of $h(x)$ from a collection of trees.
The resulting anomaly score $s(x, n)$, for which $0$ $<$ $s(x, n)$ $\leq 1$, can be utilized as follows:
\begin{itemize}
	\item The closer $s(x, n)$ is to 1 for observation $p$, the more likely $p$ is to be anomalous.
	\item Conversely, if $s(x, n)$ is significantly lower than 0.5, the observation is almost certainly non-anomalous.
\end{itemize}

An existing study of explainability of Iforest identifies the dimensions that contribute the most to the final score \cite{siddiqui2019detecting}. In contrast to an autoencoder, an isolation forest does not offer insight as to the size and sign of the deviation from normality. A more detailed explanation of Iforest is available in \cite{liu2008isolation}.

\section{Experimental Validation of the Problem}\label{sec:problemvalidation}
Above, we argued that assessing the relative normality of characteristics of a company is core to strategic decision-making and therefore is the objective of the proposed system. 
As illustrated in our review of the cognitive psychology literature, past research gives us reasons to assume that human decision-making in the strategic domain is imperfect. 
Moreover, the assumption that assessment of normality of the relevant strategic data is difficult ultimately determines the added value of this study, the type of algorithm we should use, and the evaluation strategy to be applied. If experts are bad at the task, the added value of the learning system is high, and active learning through human annotation is not possible. 

In this section, we report the setup and results of an experiment designed to test this assumption.

\subsection{Set-up}\label{subsec: exp_setup}

As outlined in Section~\ref{sec:related_work}, human experts need to be both consistent and accurate to permit ground truth labeling. Ten study subjects were selected by an HR services company as experts based on their expertise. All subjects were from the consulting division, and had either consulting, business intelligence, or director roles in the organization.

The data used in this study belongs to an HR services provider, and includes data on both employees and employers. Two datasets were composed: the first dataset (D1) consisted of $128,820$ observations of employees and included five dimensions; the second dataset (D2) consisted of $1,864$ observations of employers and included eleven dimensions. 

While data-driven outlier detection methods yield rankings efficiently, experts cannot be expected to produce a consistent ranking across a vast body of observations. Hence, for both datasets the subjects were asked to label a small subset of observations. 
First, the three methods were run on both D1 and D2.
Second, using these results, subsets were selected to (i) span the full range of normality, including observations with high, medium and low outlier rankings across methods, and (ii) ensure discrimination between methods by including a mix of observations the three methods disagreed on, i.e., ranked in very different deciles. 
Third, the ten subjects were given as much time as needed to review and label the observations as normal ($Y=0$), outlier ($Y=1$), or undecided if a subject could not decide on a label ($Y=na$). 
Furthermore, two additional indicators of aptitude of subjects were collected for both D1 and D2. Each subject was asked to score the following:
\begin{itemize}
	\item The relevance of the subject's professional experience to the labeling task on a scale of one to ten, with a score of ten meaning very relevant; and
	\item The difficulty of the labeling task on a scale of one to ten, with a score of ten meaning very difficult.
\end{itemize}

To assess whether humans indeed rely on certain heuristics when faced with complex, i.e., high-dimensional, decision-making tasks, the subjects were asked to identify the main dimensions that contributed to deciding on the label for an observation. 

Recall that a key requirement for logical decision-making, either by humans or systems, is consistency \cite{de2006frames, kahneman2013choices}. To assess the consistency of subjects, in both series of observations that were to be labeled, a number of duplicates, i.e., copies of observations, were included. 
The consistency of a subject is then evaluated as the proportion of the copied observations that were assigned the same label, 
or, for a number of subjects $s = 1,2,...,N$ and $\mathcal{D}$ duplicates,
\begin{equation}
Consistency = \sum_{i=1}^{\mathcal{D}} \frac{c_{s,i}}{\mathcal{D}},\\
\end{equation}
\begin{equation}
\text{where } c_{s,i} = \begin{cases}
1 & \text{if $s$ assigned $i$ the same label}.\\
0 & \text{if $s$ assigned $i$ a different label}.\\
\end{cases}
\end{equation}
This measure of consistency is interpreted as a proxy for proficiency at the task at hand. A higher level of inconsistency in making decisions points to irrational and non-systematic judgment.

\begin{table*}[]
	\centering
	\resizebox{\columnwidth}{!}{
	\begin{threeparttable}
		\caption{Expert Results}
		\label{tab:my-tablestats}
		\begin{tabular}{@{}llrrrrrrr@{}}
			\toprule
			&               &         &       &      &                            &  & Correlation &                                    \\ \cmidrule(l){7-9} 
			&               & Average & StDev & Min  & Max   & Consistency           & Difficulty                              & Job Relevance \\
			\midrule
			\multicolumn{1}{l|}{Employee}         & Consistency   & 71.11\% & 1.26  & 4.00 & 8.00  & 1.00                  & -0.64                                   & 0.67          \\
			\multicolumn{1}{l|}{n = 49} & Difficulty    & 6.90    & 3.11  & 2.00 & 10.00 & -0.64                 & 1.00                                    & -0.85         \\
			\multicolumn{1}{l|}{$\mathcal{D}$ = 9}         & Job relevance & 5.20    & 3.19  & 1.00 & 10.00 & 0.67                  & -0.85                                   & 1.00          \\
			\midrule
			\multicolumn{1}{l|}{Employer}         & Consistency   & 60.00\% & 1.05  & 1.00 & 5.00  & 1.00                  & 0.04                                    & 0.38          \\
			\multicolumn{1}{l|}{n = 40} & Difficulty    & 6.00    & 2.45  & 2.00 & 9.00  & 0.04                  & 1.00                                    & -0.49         \\
			\multicolumn{1}{l|}{$\mathcal{D}$ = 5}         & Job relevance & 5.60    & 2.50  & 1.00 &9.00  & 0.38                  & -0.49                                   & 1.00          \\ \bottomrule
		\end{tabular}
	\end{threeparttable}
	}
	\vspace{-6mm} 
\end{table*}

\subsection{Results}
\begin{figure}[h] 
	\centering
	\includegraphics[scale=0.35]{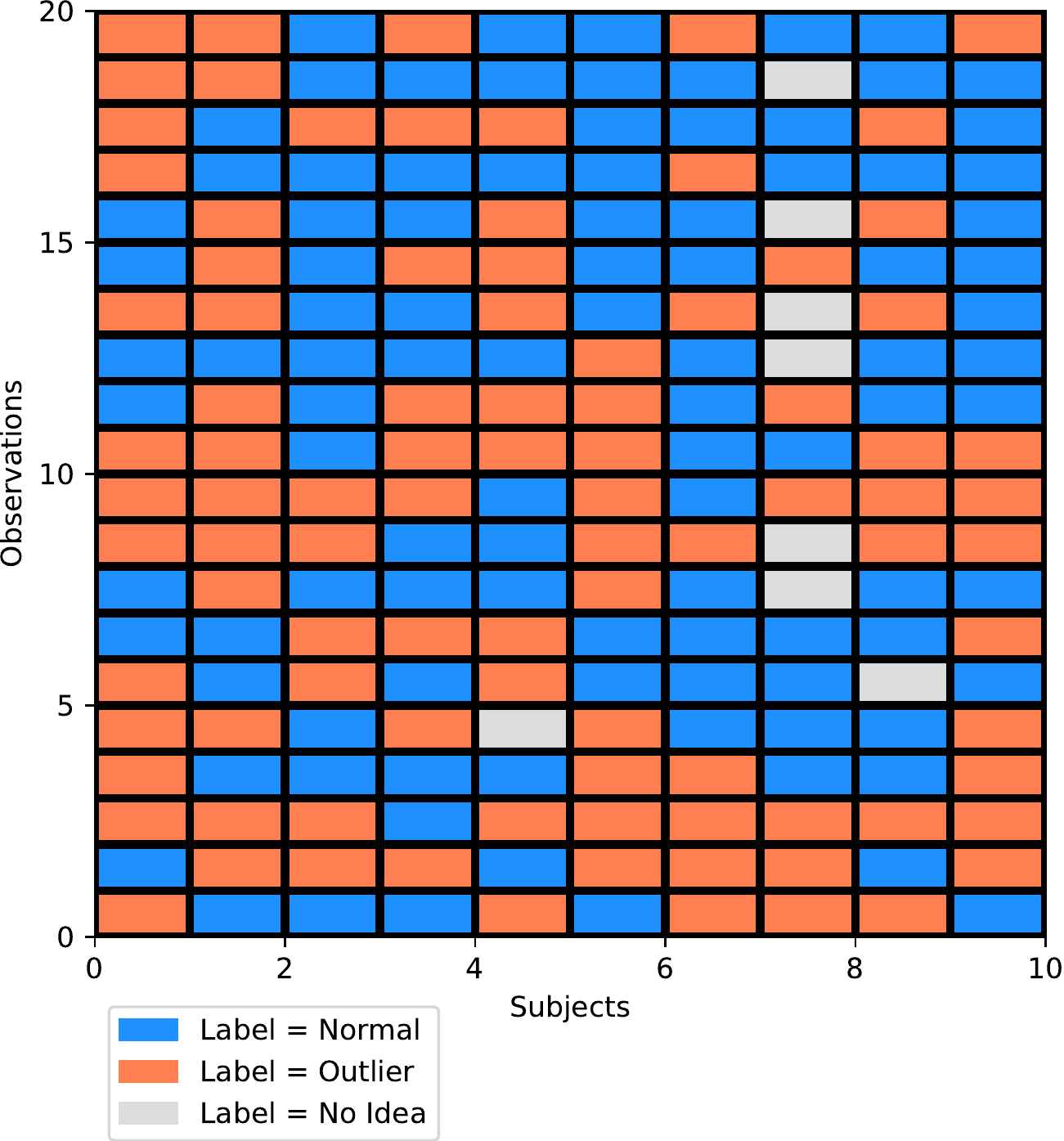}
	\caption{Expert labels for the first twenty observations of the employer dataset (D2) described in Section~\ref{sec:problemvalidation}}
	\label{fig:grid}
	\vspace*{-3mm}
\end{figure}

\begin{figure}[h] 
	\begin{center}
		\includegraphics[scale=0.44]{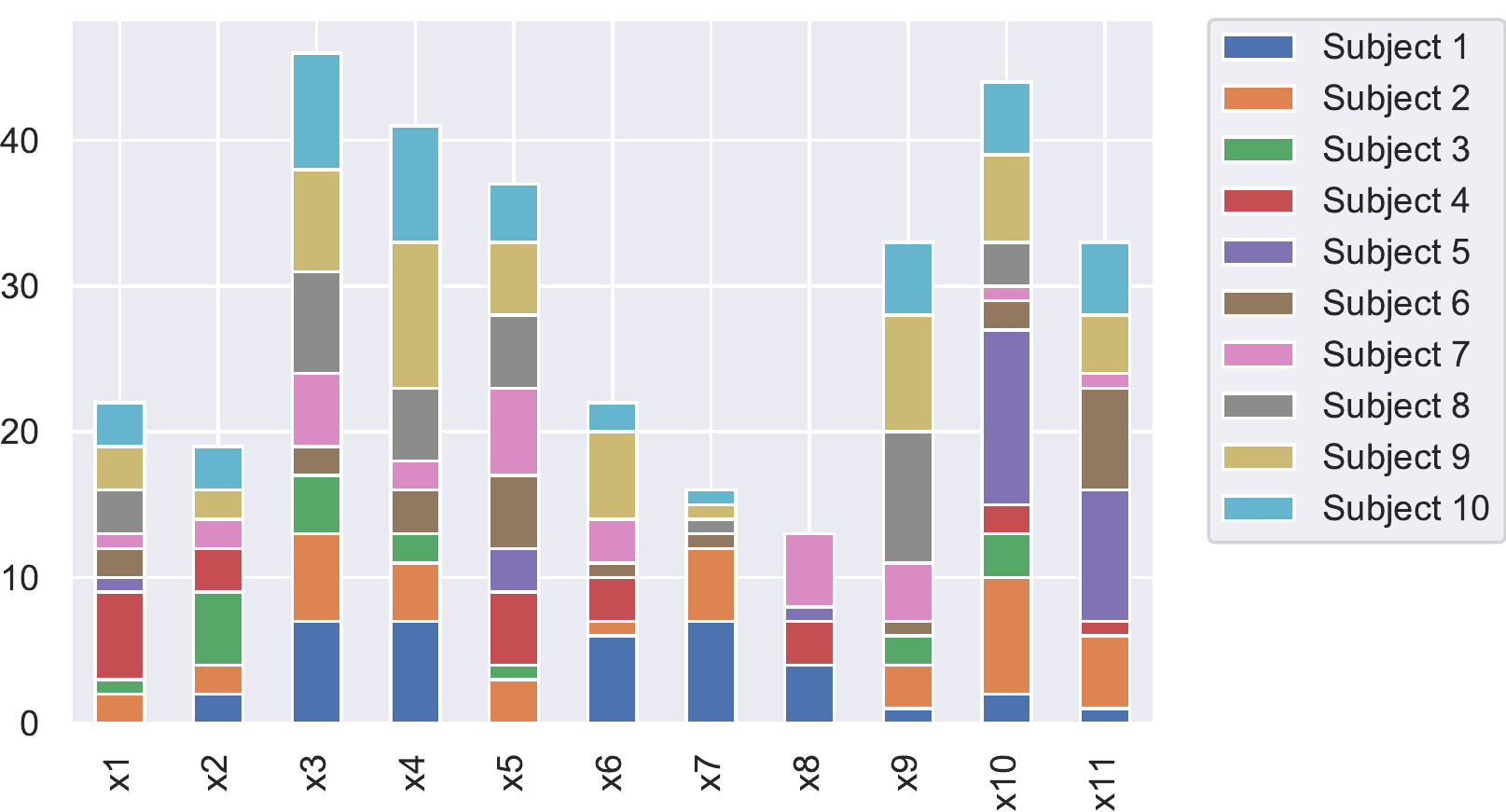}
		\caption{Distribution of the use of eleven dimensions (x1--x11) in the employer dataset (D2) by the ten experts}
		\label{fig3}
	\end{center}
	\vspace*{-3mm}
\end{figure}

For both datasets, consistency scores, indicators of aptitude, and the correlation matrix between consistency and aptitude indicators are listed in Table~\ref{tab:my-tablestats}. Four key observations can be made with respect to the results.
\begin{enumerate}
    \item First, the experts are often in disagreement with \textit{each other}, as shown in Figure~\ref{fig:grid}. The grid in this figure visualizes the expert labels for the first twenty observations of the employer dataset. The colors represent the labels, each column contains the labels assigned by a given expert, and each row visualizes the labels assigned to a given observation. Note that the experts do not unanimously agree for even a single observation on the appropriate label. Spearman rank correlation results for the judgments of individual experts are shown in Table~\ref{tab:spearmand1} and Table~\ref{tab:spearmand2} for D1 and D2, respectively.
    \item Second, the experts do not agree with \textit{themselves}. With average consistency rates of the duplicate labels of $71.11\%$ and $60.00\%$ for datasets D1 and D2, respectively, human experts are remarkably inconsistent. They appear to barely surpass random performance, characterized by a consistency rate of $50\%$. In agreement with the literature and intuitive expectations, the consistency of experts is observed to decline as complexity increases, with the consistency rate being substantially lower for the higher-dimensional dataset (namely, D2 that includes eleven dimensions). Inconsistencies are not related to a specific subset of observations that are difficult to assess, as all fourteen duplicate observations were inconsistently labeled at least once.
    \item Third, experts are observed to apply heuristics in their decision-making. Most experts focus on a relatively small number of dimensions. This can be inferred from Figure~\ref{fig3}. Moreover, experts take into account different (combinations of) dimensions in deciding on the appropriate labels. The tenth dimension is the only characteristic reported as having been used at least once by every expert. Conversely, only four experts indicated using the eighth dimension. The results also show that individual experts often combine the same dimensions, which implies heuristic decision-making.
    \item Fourth, for the employee dataset (D1), consistency is positively correlated with self-reported professional relevance, and negatively with perceived difficulty. For the employer set (D2), which includes more dimensions, the experts' self-assessment of perceived difficulty did not correlate significantly with consistency.
\end{enumerate}

While the more complex, higher-dimensional labeling task was perceived as significantly easier, performance on it was significantly worse, with a greater underestimation of difficulty for the worse performers. This result could indicate that the experts lack the meta-cognitive insight to recognize their own relative incompetence at a task.

Although these results do not provide conclusive evidence, they indicate that human experts are inconsistent decision-makers who rely on heuristic decision-making rules in complex environments. These results therefore highlight the need for granular feedback and unsupervised learning to support the problem analysis phase in strategic decision-making processes, as offered by the proposed autoencoder solution.

\section{Experimental Validation of the Solution} \label{sec:experiments}

Section~\ref{sec:problemvalidation} presented experimental evidence of the limited ability of human experts to consistently analyze complex data within their field of expertise, which is the problem we aim to address in this study. In Section~\ref{sec:experiments}, use of the autoencoder as a strategic decision support system is validated as a potential solution to this problem.

In line with the requirements defined in Section~\ref{subsec:requirements}, we identify three dimensions in validating the proposed approach: 
\begin{enumerate}[label=(\roman*)]
	\item \textbf{Outlier detection performance}: We assess the correctness of the obtained outlier score ranking, with outlier scores being the aggregated amount of deviation across all dimensions.
	\item \textbf{Dimension-level feedback}: To ensure the added value of providing granular feedback, i.e., feedback regarding size and sign of a deviation provided by the system at the level of individual dimensions, we assess the reliability and accuracy of the provided feedback.	
	\item \textbf{Synergy between the model and human assessment}: A seamless integration within the management decision-making process is vital for a successful adoption of the proposed system; here, we ensure that users correctly understand the output of the system, can use the output for practical decision-making, and do not find their personal beliefs to be in persistent conflict with the output. 
\end{enumerate}  

To validate the autoencoder-based support system across these three dimensions, we perform four experiments involving blind expert validation (Section~\ref{subsec:Blindexpertvalid}), transparent expert validation (Section~\ref{subsec:Transparantexpval}), an observed case of corrupted data (Section~\ref{subsec:Dataqual}), and synthetic observations (Section~\ref{subsec:Synthpoints}).

Table~\ref{table:valmet} summarizes the contributions of these four experiments to the validation of the system across the three dimensions identified above. The following sections will provide full details on the setup of these experiments and discuss the results. Hyperparameters and correlations are consistent with previous studies and reported in the Appendix in Table~\ref{tab:hyperparameters} and Table~\ref{tab:corrtable}.

\begin{table*}[!t]
	\centering
	\begin{adjustbox}{max width=\textwidth}
	\begin{threeparttable}
		\caption{Validation of Methodology}
		\label{table:valmet}
		\begin{tabular}{@{}lccc@{}}
			\toprule
			& \multicolumn{1}{l}{(i) Outlier detection performance} & \multicolumn{1}{l}{(ii) Synergy} & \multicolumn{1}{l}{(iii) Dimension-level feedback} \\ \midrule
			5.1. Blind expert validation & (x) & x & (x)  \\ 
			5.2. Transparent expert validation & & x  & x  \\
			5.3. Data quality  & x &  & x  \\
			5.4. Synthetic observations & x & x & x  \\ \bottomrule
		\end{tabular}
		\begin{tablenotes}
			\item[a] (x) indicates a moderate contribution.
			\item[b] x indicates a sizable contribution. 
		\end{tablenotes}
	\end{threeparttable}
	\end{adjustbox}
\end{table*}

\subsection{Blind Expert Validation}\label{subsec:Blindexpertvalid}
This first experiment aims at evaluating the accuracy of outlier scores produced by the autoencoder. Since the observations in the data are unlabeled, there is no objective ground truth that can be used for assessing the accuracy of the ranking. As argued in Section~\ref{sec:problemvalidation}, the alternative of using labels assigned by an individual human expert cannot be assumed to yield a trustworthy assessment.
As an improved alternative to using the labels of a single expert for validation, we may instead compare the assessment of the autoencoder system with that of a group of experts, which can be considered to be an ensemble classification system. An ensemble classifier benefits from accurate and diverse members \cite{dietterich2000ensemble, hansen1990neural}. Hence, we use ensemble theory to construct a weighted aggregate classifier from individual expert opinions. Every subject is considered to be a weak classifier, and it is hypothesized that their joint performance may be better, leveraging the wisdom of crowds \cite{galton1907}. 
\subsubsection{Set-up.}

To combine individual estimates, two variants of majority voting are implemented:

\textbf{Unweighted Majority Voting.}
Denote the decision of the $s^{th}$ subject (i.e., expert) by $d_{s,j} \in \{{0,1}\}$ for $s = 1, ..., S$ and $j = 1, ..., C$, where $S$ is the number of subjects, and $C$ is the number of classes, such that $d_{s,j} = 1$ for the class the subject selected, and zero otherwise. For an observation, $J_{uv}$ is the voted label, and the summation tabulates the number of votes for class $j$:
\begin{equation}
J_{uv} = argmax_{j \in \{ 0,1,2 \}} \sum_{s=1}^{S} d_{s,j}.
\end{equation}

\textbf{Weighted Majority Voting.}
Here, $w$ acts as a weighting factor for the vote. The weighted majority vote is $J_{wv}$, and the summation in this case tabulates the weighted vote for class $j$. Hence, the votes of individuals who perceive their expertise to be more relevant to the task will have larger weights in the vote.
\begin{equation}
J_{wv} = argmax_{j \in \{ 0,1,2 \}} \sum_{s=1}^{S}w _s d_{s,j},
\end{equation}
where $w = 1, ..., 10$, and $w_s$ is either the self-perceived job relevance of subject $s$ or the inverse of the self-perceived difficulty of the task of subject $s$.

The labels of individual experts, obtained in the experiment discussed in Section \ref{sec:problemvalidation} and combined using the two majority voting schemes described above, are used to assess the outlier scores of the autoencoder, LOF, and iForest by subsequently labeling five, ten, and fifteen percent of observations with the highest outlier scores as outliers. 
Afterwards, we measure the accuracy of the weighted and unweighted majority expert ensemble against this labeling. Under the assumptions that (i) the models are valid tools for outlier detection in this setting, and (ii) humans make different mistakes that can average out when combined, convergence between the labels of outlier detection methods and those of the expert ensemble is to be expected.

\subsubsection{Results.} 

Table~\ref{tab:majvores} shows the results for the unweighted and weighted expert ensembles, both when weighting with the self-reported job relevance and difficulty scores. A higher accuracy means there is a stronger match between the expert ensemble and the outlier detection method. This table demonstrates that, generally, the autoencoder attains the highest accuracy, at least in comparison with the weighted ensembles. This indicates that, among the three models, the autoencoder best matches with the weighted aggregate judgment of human experts. The absolute and percentage accuracy increases achieved by weighing the expert labels are also the highest for the autoencoder.

\begin{table*}[t]
	\centering
	\begin{adjustbox}{max width=\textwidth}
	\begin{threeparttable}
		\caption{Majority Voting Results}
		\label{tab:majvores}
		\begin{tabular}{lllllllllll}
			&               &         & AE      &         &         & Iforest     &         &         & LOF     &         \\ \toprule 
			& & 5 \%           & 10\%          & 15\%        & 5\%           & 10\%          & 15\%        & 5\%           & 10\%          & 15\%          \\ \midrule
			& Unweighted             & 0.54    & 0.56    & 0.62    & 0.56    & 0.54    & 0.59    & \textbf{{\ul 0.64}}    & 0.51    & 0.49    \\
			& JobRel\_Weight     & 0.69    & 0.72    & \textbf{{\ul 0.77}}    & 0.69    & 0.67    & 0.69    & 0.72    & 0.64    & 0.62    \\
			& Difference               & \textbf{{\ul 0.15}}    & \textbf{{\ul 0.15}}   & \textbf{{\ul 0.15}}   & 0.13    & 0.13    & 0.10    & 0.08    & 0.13    & 0.13    \\
			& \% Increase      & \textbf{{\ul 28.57\%}} & 27.27\% & 25.00\% & 22.73\% & 23.81\% & 17.39\% & 12.00\% & 25.00\% & 26.32\% \\
			& Difficulty\_Weight & 0.77    & \textbf{{\ul 0.79}}   & \textbf{{\ul 0.79}}   & 0.72    & 0.69    & 0.72    & \textbf{{\ul 0.79}}     & 0.67    & 0.64    \\
			& Difference               & \textbf{{\ul 0.23}}    & \textbf{{\ul 0.23} }    & 0.18    & 0.15    & 0.15    & 0.13    & 0.15    & 0.15    & 0.15    \\
			& \% Increase        & \textbf{{\ul 42.86\%}} & 40.91\% & 29.17\% & 27.27\% & 28.57\% & 21.74\% & 24.00\% & 30.00\% & 31.58\% \\ \bottomrule
		\end{tabular}
	\end{threeparttable}
	\end{adjustbox}
	
\end{table*}

\begin{figure}[h] 
	\begin{center}
		\includegraphics[scale=0.50]{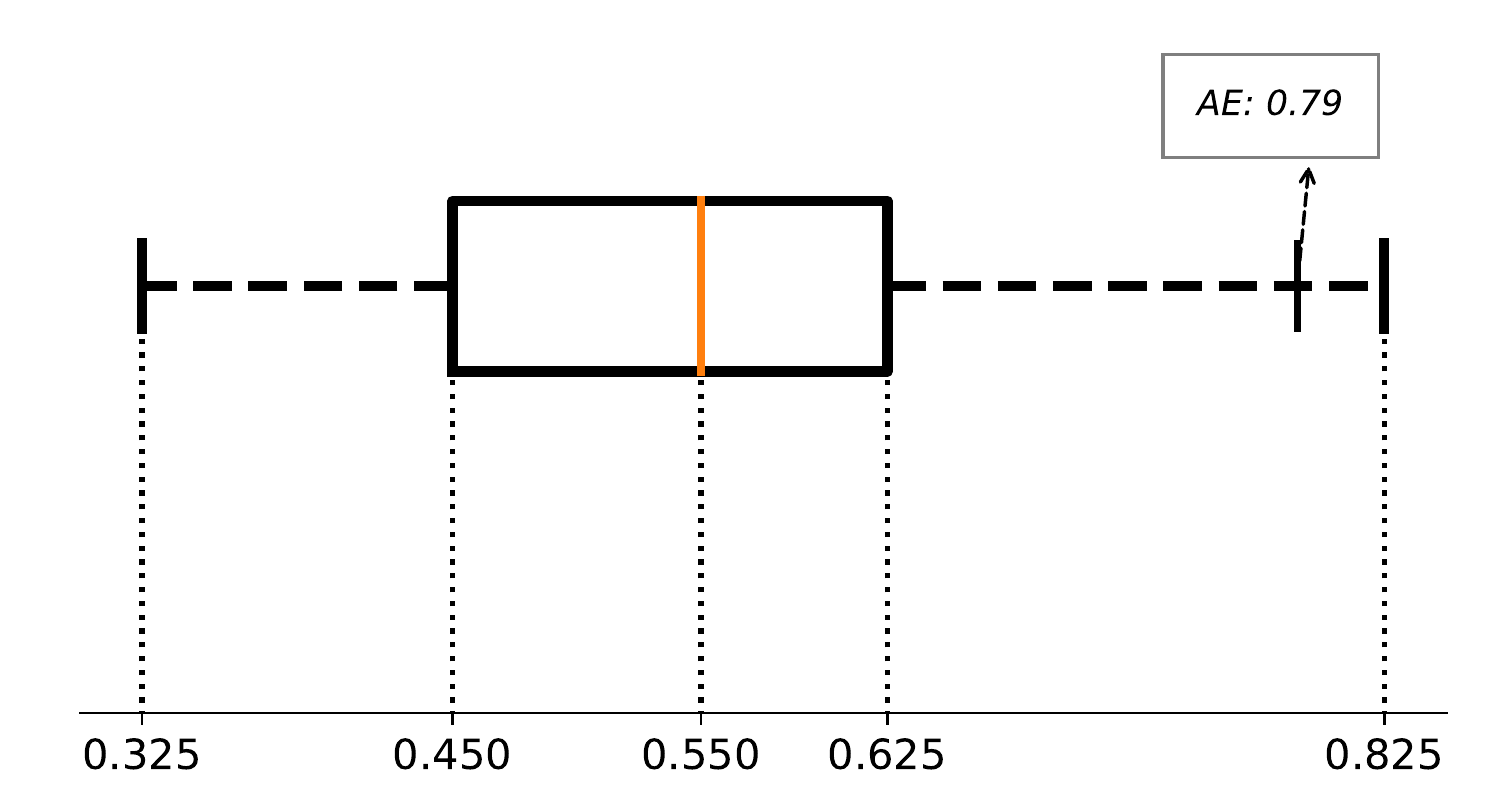}
		\caption{Individual accuracy values across all models}
		\label{boxplot}
	\end{center}
	\vspace*{-6mm}
\end{figure}

The boxplot in Figure~\ref{boxplot} represents the distribution of accuracy values of ten individual experts across the three methods (AE, LOF, Iforest) and the three cutoff values for turning outlier scores into labels ($5\%$, $10\%$, and $15\%$), thus yielding 90 data points (3x3x10). The median accuracy is barely higher than the performance of a random model. Out of these ninety combinations of cutoff values, experts and models, only one has a higher accuracy than that consistently reached by the ensemble-weighted AE. In this respect, it is remarkable that the expert-weighted ensemble stabilizes at an accuracy of just under $80\%$. This signifies that the experts make different mistakes (compared to the outlier detection labeling) and that the ensemble can leverage the individual strengths of experts and cover for individual weaknesses. It seems that the experts were relatively correct in their self-assessments, and the models are accurate, as evidenced by the accuracy increase after weighting. 

\begin{singlespacing}
\begin{table*}[t]
	\centering
	\caption{Outlier Detection Performance -- Detection Results}
	\label{tab:my-tableaccuracyresults}
	\resizebox{\columnwidth}{!}{
	\begin{tabular}{llllllllll}
		\toprule
		
		& \multicolumn{3}{l}{A}                          & \multicolumn{3}{l}{B}                          & \multicolumn{3}{l}{C}                            \\
		
		& \multicolumn{3}{l}{Data Quality}               & \multicolumn{3}{l}{Synthetic Observations}           & \multicolumn{3}{l}{Average Performance}                      \\ \midrule
		
		& 5\%           & 10\%          & 15\%        & 5\%           & 10\%          & 15\%        & 5\%           & 10\%          & 15\%          \\ \midrule
		
		AE       & \textbf{{\ul 61.80\%}} & 82.05\%       & 85.39\%     & 50.00\%       & 70.00\%       & 80.00\%     & \textbf{{\ul 55.90\%} }& \textbf{{\ul 76.01\%} }& \textbf{{\ul 82.70\%}} \\
		
		Iforest       & 60.67\%       & \textbf{{\ul 95.51\%}} & \textbf{{\ul 100.00\%}} & 10.00\%       & 20.00\%       & 40.00\%     & 35.34\%       & 57.76\%       & 70.00\%       \\
		
		LOF      & 0.00\%        & 4.49\%        & 11.24\%     & \textbf{{\ul 60.00\%}} & \textbf{{\ul 80.00\%}} & \textbf{{\ul 100.00\%} }& 30.00\%       & 42.25\%       & 55.62\%   \\ \bottomrule  
	
	\end{tabular}
	}
\end{table*}

\begin{table}[t]
	\centering
	\begin{threeparttable}
		\caption{Dimension-level Feedback -- Data Quality Set}
		\label{tab:DQ2}
		\begin{tabular}{@{}llcc@{}}
			\toprule
			Dimension & No. Obs.   & Dimension rank  & Direction \% correct   \\ \midrule
			x1             & 3   &       & 100.00\%         \\
			x2              & 5   &       & 100.00\%            \\
			x3              & 8   &       & 100.00\%           \\
			x4              & 22   &       & 100.00\%         \\
			x5              & 69  &        & 100.00\%     \\ \midrule
			All              & 89   & 85.39\%    & 100.00\%            \\ \bottomrule
		\end{tabular}
	\end{threeparttable}
	\vspace*{-5mm}
\end{table}

\end{singlespacing}

\subsection{Transparent Expert Validation} \label{subsec:Transparantexpval}
To evaluate synergy, we assess whether experts understand and agree with the output provided by the autoencoder system. 

\subsubsection{Set-up.} 
During a two-hour panel session, the group of experts were presented with the output of the autoencoder for the observations in the two datasets that the experts labeled in the previous experiment, as reported in Section~\ref{sec:problemvalidation}. The aim of the session was to gauge whether and how each expert could extract insights useful for decision-making from the output of the system. Specifically, the experts discussed the outlier score ranking as well as the granular feedback, i.e., deviations at the dimension level, provided by the autoencoder. To facilitate analysis, observations were presented using interactive visualizations that were implemented in a business intelligence software.

\subsubsection{Results.} 
The panel was able to easily interpret the results provided by the system. The panel did not object to a single assessment of the autoencoder (either at the aggregate outlier score level or at the granular dimension level). The interpretability and justifiability of the system, as confirmed by the experts, indicates synergy between experts and the model. While the autoencoder output was being studied, a data quality issue was noticed in the employee dataset (D1), highlighting synergy and yielding concrete actionable benefits of the model. Moreover, the experts proposed new applications of the system beyond the employees-and-employer dataset. Alternative employee- or employer-level datasets could be analyzed to provide specific insights on themes such as work fatigue, hiring, onboarding, etc. The versatility of the autoencoder-based approach, as recognized by the experts, indicates that the system is effective and useful in decision-making. Such versatility is a valuable property, allowing the proposed approach to be adopted as a comprehensive decision-support instrument for performing ad hoc analysis in support of any decision-making process, by merely compiling a dataset including a set of relevant dimensions.

\subsection{Data Quality} \label{subsec:Dataqual}
Data quality issues are closely related to outlyingness. As reported in the previous section, a large-impact data quality issue was discovered in the employee dataset during the transparent expert validation of the autoencoder output.
The discovered data quality issue (Section~\ref{subsec:Transparantexpval}) was fixed by in-house experts.
By comparing the pre- and post-fix versions of D1, the affected points could be reliably identified. 
Moreover, one could discern the involved dimensions as well as the direction of the effect.
For the affected points, the logical relations the variables abide by were violated. Consequently, the affected observations are sufficiently distinct to have a close affinity with the concept of an outlier. 

\subsubsection{Setup.} 
Using this data quality event to our advantage, two experiments were devised.
First, labeling the affected observations as one, and the others as zero allowed an evaluation of the detection performance of the algorithms.
Second, utilizing knowledge about the affected dimensions and the direction of the effect permitted testing of the granular feedback capabilities of AE.

To validate the dimension-level feedback of AE, we define two measures of accuracy: dimension rank accuracy, and direction accuracy:
\begin{itemize}
    \item \textbf{Dimension rank accuracy} equals 1 for an observation if the AE error is the highest in the actual affected dimension(s) and equals 0 otherwise.
    \item \textbf{Direction accuracy} of an observation is equal to 1 if for all affected dimensions, the direction is correctly represented by the sign of the difference between the observed value and the output value and is 0 otherwise.

\end{itemize}

Since the data quality issue was identified, we were able to assign ground truth labels to the affected dimensions and the direction. Using these labels, we could calculate the dimension rank and direction accuracy.

\subsubsection{Results.} 
Table~\ref{tab:my-tableaccuracyresults}A displays the data quality detection performance for the three algorithms. Iforest and AE perform well, as both have a high proportion of affected observations in the top percentiles of their respective rankings. In contrast, LOF performs poorly.

Considering the granular feedback, as shown in Table~\ref{tab:DQ2}, AE consistently recognizes the direction of the deviation (100\%) and ranks the perturbed dimension(s) the highest for 85.39\% of the observations. Interestingly, this performance does not change significantly if observations that AE did not correctly classify as affected (84.21\%) are omitted. This is particularly relevant to the extension from the top x percentile analysis to full-population decision support; even without high outlier scores, the granular feedback is accurate and valuable.

\begin{singlespacing}
\begin{table}[]
    \centering
	\begin{threeparttable}
		\caption{Dimension-level Feedback -- Synthetic Dataset}
		\label{tab:synth2}
		\begin{tabular}{@{}llll@{}}
			\toprule
			Obs. & Perturbations & Dimension rank acc. & Direction \\ \midrule
			1   & 1        & 1&correct         \\
			2   & 1&0        & correct         \\
			3   &1& 1        & correct         \\
			4   &1& 0        & correct         \\
			5   & 1&1        & correct         \\
			6   & 2&1        & correct       \\
			7   & 2&1        & correct         \\
			8   & 2&0        & correct         \\
			9   & 3& 1		&correct         \\
			10  & 3&1        & correct         \\ \midrule
			Average        &     & 70.00\%  & 100.00\%  \\ \bottomrule
		\end{tabular}
	\end{threeparttable}
\end{table}
\end{singlespacing}

\subsection{Synthetic observations} \label{subsec:Synthpoints}

Due to the instability of outlier detection algorithms reported in the literature across domains \cite{campos16}, examining performance on the data quality dataset is insufficient for evaluating performance in general. Therefore, we adopt the approach proposed and applied in \cite{das2007detecting, nychis2008empirical} and inject synthetic outliers into the dataset.

\subsubsection{Setup.}
Observations in the employer dataset that were evaluated as \emph{non-outlying} by three outlier detection methods were selected. Next, perturbations to these observations were devised by a panel of four experts to achieve impossibility, illogicality, or implausibility beyond a reasonable doubt with a minimum amount of perturbation. As such, variance between the methods' rankings is ensured, making it possible to discern the best-performing method. 

The synthetic observations were varied across the data plane with five unidimensional, three two-dimensional, and two three-dimensional perturbations. After their inception, these perturbed observations were added to the full dataset, and outlier detection models were retrained. To evaluate the dimension-level feedback, we use the same measures as reported in Section~\ref{subsec:Dataqual}. In this experiment, we can observe the ground truth, label accordingly, and evaluate the accuracy.

\subsubsection{Results.} 
Table~\ref{tab:my-tableaccuracyresults}B shows that AE performs well. Additionally, and in contrast with Table~\ref{tab:my-tableaccuracyresults}A, LOF reports great results, with perfect discrimination at 15\% cutoff. Iforest, however, performs poorly. The results for the dimension rank accuracy and the directional feedback are displayed in Table~\ref{tab:synth2}. For 70.00\% of the perturbed observations, AE correctly ranks all perturbed dimension(s). Furthermore, the autoencoder obtains the correct direction of the perturbation in all dimensions for every observation.

\subsection{Discussion} \label{sec:discussion}
The results of various experiments illustrate that investigating and devising methods to support human decision-making is a worthwhile effort. Especially if feedback on own performance is fuzzy, human biases can have a significant impact on decision-making. This also applies to seasoned experts with extensive professional experience. Interestingly, biases differ among individual experts and therefore may be countered by leveraging ensembles of experts, i.e., the wisdom of crowds, specifically if different levels of expertise are acknowledged. As Table~\ref{tab:majvores} and Figure~\ref{boxplot} show, the accuracy of the ensemble of experts eclipses average individual expert performance.

We validated an autoencoder-based approach as a solution designed to counter this problem and to provide decision support on three levels:

\begin{itemize}
    \item \textbf{Outlier detection performance.} Validation results using experts, data quality and synthetic observations reported in Tables~\ref{tab:my-tableaccuracyresults}A, \ref{tab:my-tableaccuracyresults}B and \ref{tab:majvores} indicate a strong performance of the autoencoder in detecting outliers in various experiments. Tables~\ref{tab:my-tableaccuracyresults}C and \ref{tab:majvores} illustrate performance for various settings and show that the autoencoder significantly outperforms two other state-of-the-art algorithms assessed in the experiments. The instability of results due to data quality and synthetic observations' settings confirms earlier results reported in \cite{campos16}, who reported instability of methods when comparing performance for different outlier detection settings. We observe this phenomenon for two datasets in the same setting. A desirable solution should therefore generalize well across semantic definitions of outliers without requiring significant hyperparameter tuning. Moreover, excessive tuning to a specific semantic definition may prevent the model from identifying interesting semantically varying patterns. Tuning on already discovered data quality issues seems especially inappropriate. Based on the results of the conducted experiments, we conclude that the autoencoder generalizes well across settings.
    \item \textbf{Synergy.} The autoencoder is shown to be highly synergistic with human decision-making processes due to (i) strongly correlating with joint weighted human decision-making, (ii) being unanimously accepted during a two-hour panel discussion that explored the insights provided by the approach, and (iii) matching the semantic definition of outliers on synthetic observations. The experts in the panel were able to interpret and explain the results, placing them in a richer context than that the model had direct access to through the input data.
    \item \textbf{Granular feedback at the dimension level.} The autoencoder is a powerful tool for discerning the rank and deviation direction of the main dimensions contributing to abnormality. Traditional outlier detection methods do not offer such granular feedback. Moreover, the autoencoder achieves perfect accuracy in assessing the direction of the deviation in our experiments (Tables~\ref{tab:DQ2} and~\ref{tab:synth2}). An interesting implication is that the dimension-level feedback seems remarkably stable even for low-ranked observations. This supports the idea of adopting unsupervised outlier detection methods for obtaining actionable information beyond a small set of top-ranked observations with high aggregate outlier scores.

\end{itemize}

\section{Conclusions}\label{sec:conclusion}

In this paper, we propose an unsupervised learning approach to support strategic decision-making by adopting the autoencoder, a powerful artificial neural network-based method that can provide detailed insights in regard to large and small deviations from what is expected. Such deviations relative to relevant peers support decision-making, providing feedback on the ``as-is'' situation and the direction towards an improved ``to-be'' situation.

To validate the proposed approach, a unique dataset was obtained from a European HR services provider, including information on a large set of employees and employers. Using a panel of ten experts, we observe that, as a first contribution to this domain, human experts are inconsistent and non-comparable in their judgments.
This finding strongly motivates the need for support in the first stage of the business decision-making process, i.e., the analysis of business problems.

To this end, we investigate the detection performance, synergy, and granular feedback of our autoencoder-based solution.
We acknowledge that in this setting, there is no single guaranteed evaluation method for assessing the performance and use of the proposed method. In the absence of a generally accepted evaluation procedure, we devise and perform four experiments for validation using (i) transparent expert validation, (ii) blind expert validation, (iii) data quality classification, and (iv) generation of synthetic observations.

The results of these experiments indicate that the proposed autoencoder method meets business users' requirements in terms of outlier detection performance, synergy, and dimension-level feedback. Moreover, the method is versatile and can be adopted to support decision-making across various management areas by compiling appropriate datasets.

Unsupervised learning for decision support is an underexplored research area. Decision support systems that interconnect humans and machines are urgently needed to unlock the potential of big data for optimizing strategic 
decision-making. Several challenges remain: \\
(i) A framework for objective and trustworthy validation of analytical models in an unsupervised setting is missing;\\
(ii) Models need to be more robust, reducing the risk of failure modes;\\
(iii) Developing a system to provide strategic decision support is a challenge to data scientists since the development of a system that aligns with high-level strategy requires a higher level of business understanding than development of traditional decision support systems, e.g., a customer churn prediction model, and\\
(iv) A lack of familiarity with unsupervised learning methods may hamper swift industry adoption.

Along with challenges, unsupervised decision support offers exciting possibilities for future research. Possible areas for further development include the following: \\
(i) The incorporation of a temporal dimension to capture and describe the time-varying nature of the data distribution;\\
(ii) The demonstration and prediction of causal effects of actions with regard to their abnormality profile;\\
(iii) The extension of other unsupervised algorithms to deliver granular population-wide decision support;\\
(iv) The investigation of the generalization capacity of various algorithms across different semantic definitions of normality; and\\
(v) A pragmatic alternative offered by our approach to the 
bandit model literature proposing fully autonomous decision systems \cite{nipsjeroen} that may offer opportunities for extending the proposed approach that are yet to be explored.

\bibliography{bibliography.bib}

\appendix
\section{Expert Rank Correlation}

\begin{table}[h]
	\centering
	\resizebox{\columnwidth}{!}{
	\begin{threeparttable}
		\caption{Expert Spearman Rank Correlation D1}  
		\label{tab:spearmand1} 
            \begin{tabular}{lrrrrrrrrrr}
            \toprule
            {} &  Expert 1 &  Expert 2 &  Expert 3 &  Expert 4 &  Expert 5 &  Expert 6 &  Expert 7 &  Expert 8 &  Expert 9 &  Expert 10 \\
            \midrule
            Expert 1  &  1.000 &  0.605 &  0.675 &  0.381 &  0.397 &  0.418 &  0.468 &  0.321 &  0.200 &   0.376 \\
            Expert 2  &  0.605 &  1.000 &  0.494 &  0.245 &  0.779 &  0.731 &  0.677 &  0.704 &  0.261 &   0.602 \\
            Expert 3  &  0.675 &  0.494 &  1.000 &  0.322 &  0.387 &  0.281 &  0.260 &  0.327 &  0.055 &   0.245 \\
            Expert 4  &  0.381 &  0.245 &  0.322 &  1.000 &  0.205 & -0.001 &  0.074 &  0.162 &  0.187 &   0.404 \\
            Expert 5  &  0.397 &  0.779 &  0.387 &  0.205 &  1.000 &  0.684 &  0.691 &  0.698 &  0.333 &   0.443 \\
            Expert 6  &  0.418 &  0.731 &  0.281 & -0.001 &  0.684 &  1.000 &  0.779 &  0.534 &  0.246 &   0.333 \\
            Expert 7  &  0.468 &  0.677 &  0.260 &  0.074 &  0.691 &  0.779 &  1.000 &  0.643 &  0.263 &   0.446 \\
            Expert 8  &  0.321 &  0.704 &  0.327 &  0.162 &  0.698 &  0.534 &  0.643 &  1.000 &  0.432 &   0.641 \\
            Expert 9  &  0.200 &  0.261 &  0.055 &  0.187 &  0.333 &  0.246 &  0.263 &  0.432 &  1.000 &   0.344 \\
            Expert 10 &  0.376 &  0.602 &  0.245 &  0.404 &  0.443 &  0.333 &  0.446 &  0.641 &  0.344 &   1.000 \\
            \bottomrule
            \end{tabular}

	\end{threeparttable}
	}
\end{table}

\begin{table}[h]
	\centering
	\resizebox{\columnwidth}{!}{
	\begin{threeparttable}
		\caption{Expert Spearman Rank Correlation D2}  
		\label{tab:spearmand2} 
            \begin{tabular}{lrrrrrrrrrr}
            \toprule
            {} &  Expert 1 &  Expert 2 &  Expert 3 &  Expert 4 &  Expert 5 &  Expert 6 &  Expert 7 &  Expert 8 &  Expert 9 &  Expert 10 \\
            \midrule
            Expert 1  &  1.000 &  0.053 &  0.221 & -0.234 & -0.082 & -0.016 &  0.305 &  0.004 &  0.430 &   0.219 \\
            Expert 2  &  0.053 &  1.000 &  0.218 &  0.258 &  0.261 &  0.402 &  0.100 &  0.519 &  0.027 &   0.277 \\
            Expert 3  &  0.221 &  0.218 &  1.000 &  0.169 &  0.130 &  0.175 &  0.065 &  0.150 &  0.224 &   0.381 \\
            Expert 4  & -0.234 &  0.258 &  0.169 &  1.000 &  0.380 &  0.129 & -0.181 &  0.061 & -0.109 &   0.424 \\
            Expert 5  & -0.082 &  0.261 &  0.130 &  0.380 &  1.000 &  0.032 & -0.177 &  0.205 &  0.251 &   0.023 \\
            Expert 6  & -0.016 &  0.402 &  0.175 &  0.129 &  0.032 &  1.000 &  0.090 &  0.471 & -0.067 &   0.373 \\
            Expert 7  &  0.305 &  0.100 &  0.065 & -0.181 & -0.177 &  0.090 &  1.000 &  0.168 &  0.034 &   0.428 \\
            Expert 8  &  0.004 &  0.519 &  0.150 &  0.061 &  0.205 &  0.471 &  0.168 &  1.000 &  0.177 &   0.201 \\
            Expert 9  &  0.430 &  0.027 &  0.224 & -0.109 &  0.251 & -0.067 &  0.034 &  0.177 &  1.000 &  -0.042 \\
            Expert 10 &  0.219 &  0.277 &  0.381 &  0.424 &  0.023 &  0.373 &  0.428 &  0.201 & -0.042 &   1.000 \\
            \bottomrule
            \end{tabular}

	\end{threeparttable}
	}
\end{table}

\newpage
\section{Implementation details}

\begin{table}[htbp]
	\centering
	\begin{threeparttable}
		\caption{Implementations and parameter selection}  
		\label{tab:hyperparameters} 
		\begin{tabular}{@{}ll@{}}
			\toprule
			\textbf{Model}   & \textbf{Parameters} \{employee, employer\}                                                                                                                                                        \\ \midrule
			AE      & \begin{tabular}[c]{@{}l@{}}Hidden layers: 3,8\\ Encoding dimensions: 4,7\\ Activation function: `SELU'\\Loss = `MSE' \\Optimizer: Adam\\ Learning rate: 9.5e-3\end{tabular} \\
			Iforest & \textit{Contamination} = 0.5\\ &Distance: Minkowski with \textit{p = 2}                                                                                                                                   \\
			LOF     & \textit{k}=max(\textit{n}*0.1,50)                                                                                                                                         \\ \bottomrule
		\end{tabular}
	\end{threeparttable}
\end{table}

\begin{table}[htbp]
    \caption{Correlation Methods}
    \label{tab:corrtable}
    \centering
    \begin{tabular}{@{}lllllllllll@{}}
    \toprule
    Correlations       &           &            & AE         &            &            & Iforest        &            &            & LOF        &            \\ \midrule
                &           & 5\%        & 10\%       & 15\%       & 5\%        & 10\%       & 15\%       & 5\%        & 10\%       & 15\%       \\
                & 5\%       & 1.00       & 0.83       & 0.67       & 0.57       & 0.47       & 0.39       & 0.55       & 0.64       & 0.61       \\
    AE          & 10\%      & 0.83       & 1.00       & 0.81       & 0.46       & 0.46       & 0.45       & 0.77       & 0.77       & 0.73       \\
                & 15\%      & 0.67       & 0.81       & 1.00       & 0.49       & 0.44       & 0.59       & 0.60       & 0.75       & 0.70       \\
                & 5\%       & 0.57       & 0.46       & 0.49       & 1.00       & 0.73       & 0.54       & 0.53       & 0.46       & 0.44       \\
    Iforest         & 10\%      & 0.47       & 0.46       & 0.44       & 0,73       & 1.00       & 0.75       & 0.42       & 0.41       & 0.38       \\
                & 15\%      & 0.39       & 0.45       & 0.59       & 0.54       & 0.75       & 1.00       & 0.45       & 0.34       & 0.30       \\
                & 5\%       & 0.55       & 0.77       & 0.60       & 0.53       & 0.42       & 0.45       & 1.00       & 0.68       & 0.65       \\
    LOF         & 10\%      & 0.64       & 0.77       & 0.75       & 0.46       & 0.41       & 0.34       & 0.68       & 1.00       & 0.95       \\
                & 15\%      & 0.61       & 0.73       & 0.70       & 0.44       & 0.38       & 0.30       & 0.65       & 0.95       & 1.00       \\ \bottomrule

    \end{tabular}
\end{table}
\end{document}